\documentclass{ecai}

\usepackage{latexsym}
\usepackage{amssymb}
\usepackage{amsmath}
\usepackage{amsthm}
\usepackage{booktabs}
\usepackage{enumitem}
\usepackage{graphicx}
\usepackage{color}

\usepackage{times}
\usepackage{caption}
\usepackage{subcaption}
\usepackage{algorithm}
\usepackage[noend]{algpseudocode}
\usepackage{amsfonts}
\usepackage{amsmath}
\usepackage{newfloat}
\usepackage{listings}

\usepackage{todonotes}
\usepackage{cleveref}

\newtheorem{definition}{Definition}

\DeclareMathAlphabet{\mathpzc}{OT1}{pzc}{m}{it}

\newcommand{\tuple}[1]{\ensuremath{\langle #1 \rangle}}

\newcommand{\myparagraph}[1]{\vskip2pt\noindent\textbf{#1.}}

\algrenewcommand\alglinenumber[1]{\scriptsize #1}

\algrenewcommand\algorithmicindent{1.3em}%

\errorcontextlines\maxdimen

\makeatletter
    \newcommand*{\algrule}[1][\algorithmicindent]{\makebox[#1][l]{\hspace*{.4em}\thealgruleextra\vrule height \thealgruleheight depth \thealgruledepth}}%
\newcommand*{\thealgruleextra}{}
\newcommand*{\thealgruleheight}{.8\baselineskip}
\newcommand*{\thealgruledepth}{.25\baselineskip}

\newcount\ALG@printindent@tempcnta
\def\ALG@printindent{%
    \ifnum \theALG@nested>0
        \ifx\ALG@text\ALG@x@notext
        \else
            \unskip
            \addvspace{-1pt}
            \ALG@printindent@tempcnta=1
            \loop
                \algrule[\csname ALG@ind@\the\ALG@printindent@tempcnta\endcsname]%
                \advance \ALG@printindent@tempcnta 1
            \ifnum \ALG@printindent@tempcnta<\numexpr\theALG@nested+1\relax
            \repeat
        \fi
    \fi
    }%
\usepackage{etoolbox}
\patchcmd{\ALG@doentity}{\noindent\hskip\ALG@tlm}{\ALG@printindent}{}{\errmessage{failed to patch}}
\makeatother

\newbox\statebox
\newcommand{\myState}[1]{%
    \setbox\statebox=\vbox{#1}%
    \edef\thealgruleheight{\dimexpr \the\ht\statebox+.5pt\relax}%
    \edef\thealgruledepth{\dimexpr \the\dp\statebox+.5pt\relax}%
    \ifdim\thealgruleheight<.5\baselineskip
        \def\thealgruleheight{\dimexpr .5\baselineskip+.5pt\relax}%
    \fi
    \ifdim\thealgruledepth<.1\baselineskip
        \def\thealgruledepth{\dimexpr .1\baselineskip+.5pt\relax}%
    \fi
    \State #1%
    \def\thealgruleheight{\dimexpr .5\baselineskip+.5pt\relax}%
    \def\thealgruledepth{\dimexpr .1\baselineskip+.5pt\relax}%
}

\algnewcommand\algorithmicforeach{\textbf{for each}}
\algdef{S}[FOR]{ForEach}[1]{\algorithmicforeach\ #1\ \algorithmicdo}

\makeatletter
\newcommand\footnoteref[1]{\protected@xdef\@thefnmark{\ref{#1}}\@footnotemark}
\makeatother

\begin{document}


\begin{frontmatter}
\title{A Meta-Engine Framework for Interleaved Task and Motion Planning using Topological Refinements}
\author[A]{\fnms{Elisa}~\snm{Tosello}\thanks{Corresponding Author. Email: etosello@fbk.eu}}
\author[A]{\fnms{Alessandro}~\snm{Valentini}}
\author[A]{\fnms{Andrea}~\snm{Micheli}}
\address[A]{Fondazione Bruno Kessler, Trento, Italy}
\begin{abstract}
Task And Motion Planning (TAMP) is the problem of finding a solution to an automated planning problem that includes discrete actions executable by low-level continuous motions. This field is gaining increasing interest within the robotics community, as it significantly enhances robot's autonomy in real-world applications. Many solutions and formulations exist, but no clear standard representation has emerged. In this paper, we propose a general and open-source framework for modeling and benchmarking TAMP problems. Moreover, we introduce an innovative meta-technique to solve TAMP problems involving moving agents and multiple task-state-dependent obstacles. This approach enables using any off-the-shelf task planner and motion planner while leveraging a geometric analysis of the motion planner's search space to prune the task planner's exploration, enhancing its efficiency. We also show how to specialize this meta-engine for the case of an incremental SMT-based planner. We demonstrate the effectiveness of our approach across benchmark problems of increasing complexity, where robots must navigate environments with movable obstacles. Finally, we integrate state-of-the-art TAMP algorithms into our framework and compare their performance with our achievements.

\end{abstract}

\end{frontmatter}


\section{Introduction}\label{sec:introduction}
Task And Motion Planning (TAMP) is the problem of finding high-level plans to accomplish assigned tasks (task planning), as well as the motions needed to execute these plans (motion planning). 
Consider a warehouse robot collecting items and placing them in bins for shipment. 
At the task level, it determines the sequence of actions needed, such as collecting items and navigating. At the motion level, it plans the movements considering obstacles. Merely sequencing task and motion planning may lead to ineffective solutions, with the robot possibly moving directly toward the goal, ignoring obstacles. In contrast, integrating these components effectively allows the robot's plan to adapt dynamically. For instance, if a pallet blocks an aisle, the robot will try to move it before proceeding further.

A wide range of proposed solutions and formulations exist, but no clear standard representation has emerged~\cite{8411475}.
In this paper, we propose a formalization and implementation for modeling TAMP problems related to navigation tasks involving multiple movable objects, which remains independent of specific planners and languages. Additionally, we offer an open-source modeling tool, built within the open-source Unified Planning (UP) library\footnote{\label{up}Available at \url{https://github.com/aiplan4eu/unified-planning}}, that facilitates seamless integration of setups and planners for evaluation and comparison. As evidence of our approach, we provide an exhaustive benchmarks suite aligned with existing TAMP evaluation criteria~\cite{8411475}, such as handling infeasible task actions, managing large task spaces, and balancing the trade-off between task complexity and motion execution.

Furthermore, we devise and integrate into our overall framework a planning technique tailored to this class of problems that allows to combine off-the-shelf automated planners with off-the-shelf motion planners. 
We exploit the \textit{Meta-Engine} feature of the UP library to instantiate our framework with any task planner available through the library. Then, we use the Open Motion Planning Library (OMPL)~\cite{ompl} to plan motions (but any other solver could be exploited). Our approach fits into the category of interleaved TAMP~\cite{Garrett2021, Dantam2020}: a Benders Decomposition~\cite{BENDERS1962/63} of the TAMP problem where the automated task planner decides a candidate plan disregarding the motion constraints. Then, the motion planner tries to refine the plan by adding the motion details. If it fails, it analyzes the reason for the failure and derives an explanation that the task planner can use to prune its search for new plans. In this sense, the core of our approach is what we call \textit{topological refinement}: we approximate the area explored by the motion planner, derive the encountered obstacles, and exploit them to formulate new constraints that we add at the task level. This refinement allows us to prune entire symbolic space regions rather than just the immediate unrealizable action, as typically done in traditional TAMP approaches.

One drawback of using an off-the-shelf automated planner is the need to restart the task planning search every time the framework learns a new constraint. Hence, we also present a simple but effective algorithm, called \textsc{Tampest} (Task And Motion Planning by Encoding into Satisfiability Testing), for task planning based on the Satisfiability Modulo Theory (SMT)~\cite{smt} framework that can avoid restarts by exploiting the incrementality feature of modern SMT solvers.

Finally, we integrated \textsc{PDDLStream}~\cite{PDDLStream}, a solver increasingly used in TAMP, into our framework, making it one of the solvers supported by the UP library. Given our TAMP formulation, which is accepted as input by any UP-supported solver, we automatically convert it to the format supported by PDDLStream, so it can be used to solve TAMP problems without the need for customization.
This integration demonstrates our ability to completely separate problem formulation from the solving algorithm, empowering users to compare various solvers using identical problem formulations and input data.
We include a thorough experimental assessment, comparing various task planners, sampling-based motion planners, and benchmarking against \textsc{PDDLStream}. We show \textsc{Tampest}'s effectiveness and efficiency, particularly with topological refinements.

The paper is organized as follows. 
Section~\ref{sec:problem_statement} formalizes our TAMP problem, and Section~\ref{sec:meta_engine} introduces our meta-engine approach, with its SMT-based specialization in Section~\ref{sec:snt-framework}. 
Section~\ref{sec:modeling_and_benchnmarking} details our benchmarks, Section~\ref{sec:state-of-the-art} reviews related work, and Section~\ref{sec:experimental_evaluation} discusses experiments and results. Finally, in Section~\ref{sec:conclusions} we draw our conclusions.

\section{Problem Statement}\label{sec:problem_statement}
\begin{figure*}[t]
     \centering
     \begin{subfigure}[b]{0.242\textwidth}
         \centering
         \includegraphics[height=92pt]{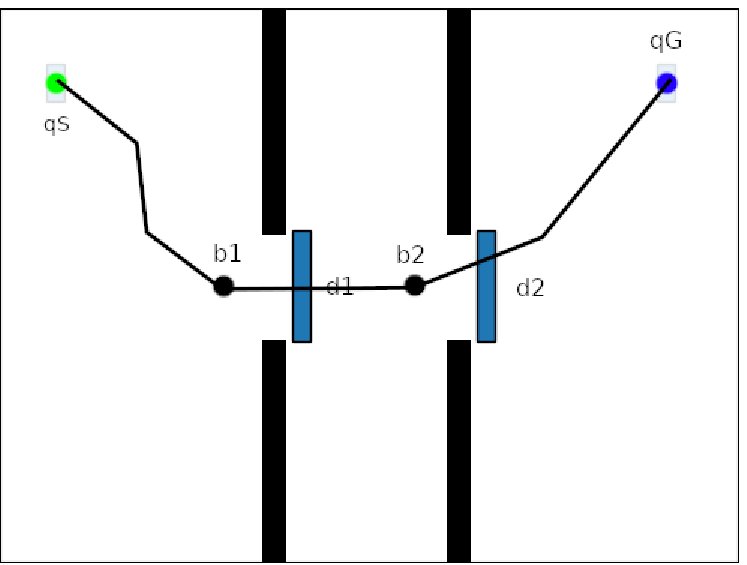}
         \caption{}
         \label{subfig:example_0}
     \end{subfigure}
     \hfill
     \begin{subfigure}[b]{0.242\textwidth}
         \centering
         \includegraphics[height=92pt]{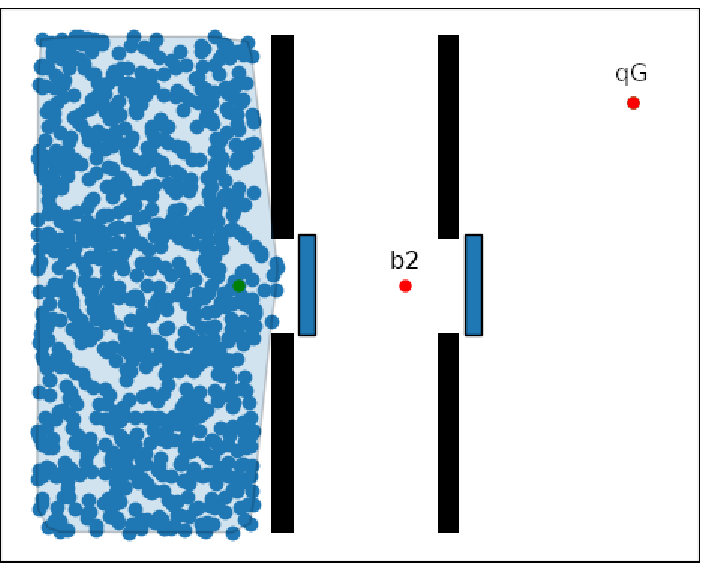}
         \caption{}
         \label{subfig:example_1}
     \end{subfigure}
     \hfill
     \begin{subfigure}[b]{0.242\textwidth}
         \centering
         \includegraphics[height=92pt]{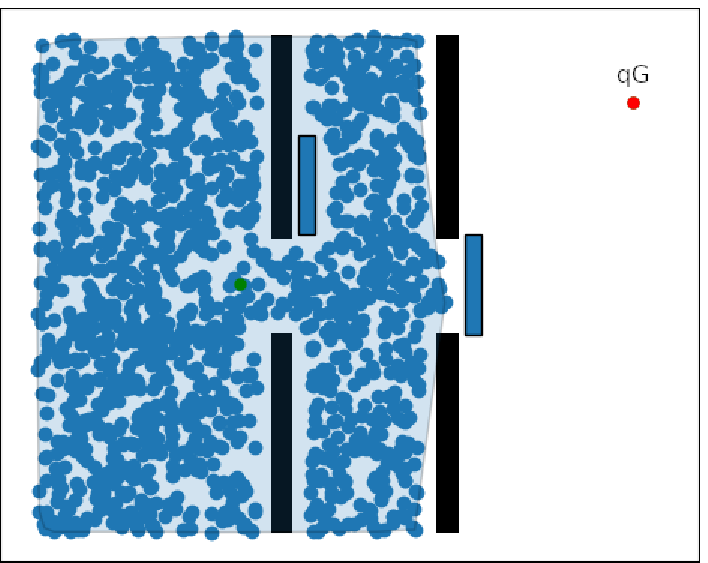}
         \caption{}
         \label{subfig:example_2}
     \end{subfigure}
     \hfill
     \begin{subfigure}[b]{0.242\textwidth}
         \centering
         \includegraphics[height=92pt]{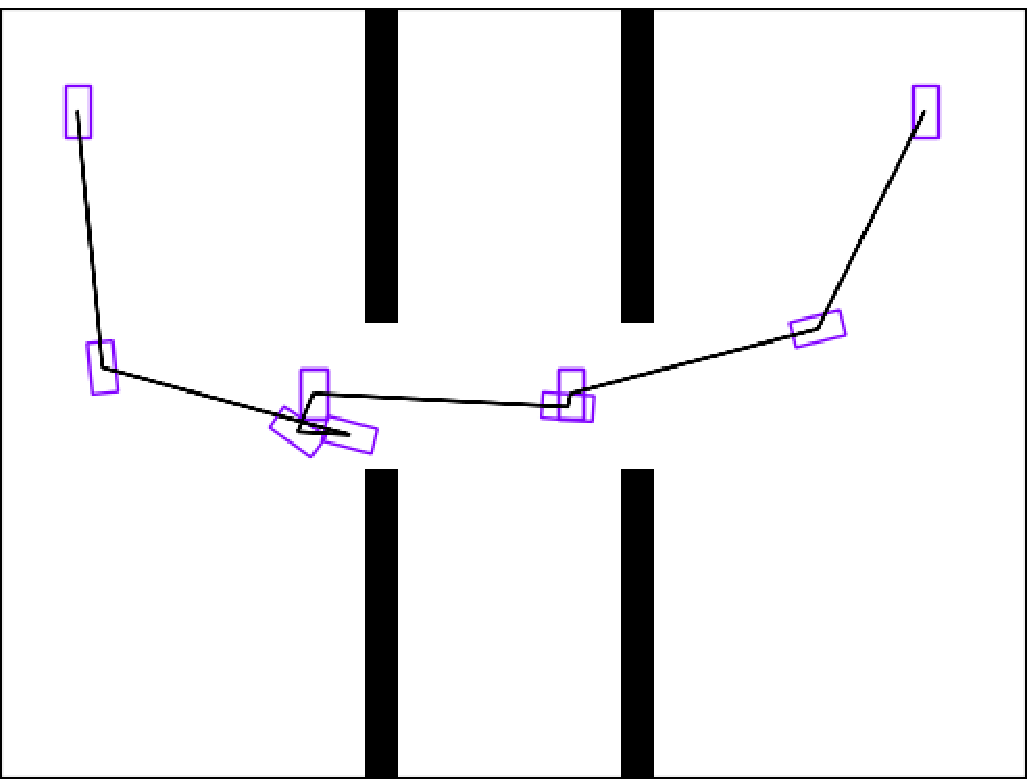}
         \caption{}
         \label{subfig:example_3}
     \end{subfigure}
     \vspace{0.22cm}
    \caption{The \textit{Doors} domain. \subref{subfig:example_0} A robot $r$ has to move from $q_S$ to $q_G$, passing through doors \{$d_1, d_2$\}, initially closed. Each door opens when its button is pushed. \subref{subfig:example_1} Initially, $r$ tries to reach $q_G$ but finds $d_1$ closed, shown by blue dots indicating sampled configurations.\subref{subfig:example_2}  After opening $d_1$, it tries to reach $q_G$ again but finds $d_2$ closed. \subref{subfig:example_3} Only after opening both $d_1$ and $d_2$, $r$ finds a collision-free path from $q_S$ to $q_G$.}
    \vspace{0.2cm}
    \label{fig:doors_domain}
\end{figure*}

In this Section, we formalize a TAMP problem with mobile agents moving within a workspace populated by task-dependent obstacles.
As a motivating example, consider a robot tasked to navigate an office environment with multiple sliding doors controlled by button presses (see Figure~\ref{fig:doors_domain}). To reach its destination, the robot needs to find a sequence of actions to move and open doors. Simultaneously, it must physically execute these actions, which means finding motion primitives ensuring collision-free movement. Upon pressing the button, it must be aware of the change in door configuration so it can pass through and reach the assigned target.
A formal definition of this class of problems follows.

\begin{definition}\label{def:tamp}
A (ground) \textbf{Task And Motion Planning problem} is a tuple $\psi$ = $\langle\mathpzc{R}, \mathpzc{W}, \mathpzc{C}, \mathpzc{U}, \mathpzc{V}, \mathpzc{I}, \mathpzc{A}, \mathpzc{G}\rangle$ such that:
\begin{itemize}
    \item $\mathpzc{R}$ is a set of mobile agents, where each agent $r$ is characterized by a certain geometric model.

    \item $\mathpzc{W} \subseteq \mathbb{R}^N$ (N = 2 or N = 3) is the workspace, that is the physical volume of all end point positions reachable by the robots in $\mathpzc{R}$. We define $\mathpzc{W}_{f}$ as the subset of $\mathpzc{W}$ free from fixed obstacles.

    \item $\mathpzc{U}$ is a map that assigns to each agent $r \in \mathpzc{R}$ a motion model $\mathpzc{U}_r$, that is a mathematical representation of the kinematic and dynamic laws that allows the agent to evolve within $\mathpzc{W}$.

    \item $\mathpzc{C}$ is the configuration space, where $\mathpzc{C}_r \subseteq \mathpzc{C}$ is that subset of $\mathpzc{C}$ that represents the joint configurations that $r \in \mathpzc{R}$ may assume given its motion model. 
    In this context, $occ(r, q) \subseteq \mathpzc{W}_{f}$ is the set of points in $\mathpzc{W}_{f}$ occupied by $r$ when in configuration $q \in \mathpzc{C}_r$.

    \item $\mathpzc{V}$ = $\{f_1, .., f_k\}$ is a finite set of variables (or fluents) $f \in \mathpzc{V}$, each with a finite or infinite domain $Dom(f)$.

    \item $\mathpzc{I}$ is the initial task state, which assigns a value $\mathpzc{I}(f) \in Dom(f)$ to each $f \in V$.

    \item $\mathpzc{A}$ is a set of actions $a = \langle \mathpzc{P}, \mathpzc{E}, \mathpzc{M} \rangle$ such that:
    \begin{itemize}
        \item $\mathpzc{P}$ is a set of preconditions \textit{pre} $\in \mathpzc{P}$, with \textit{pre} a Boolean combination of atoms $f = v$, with $f \in \mathpzc{V}$ and $v \in Dom(f)$.
        \item $\mathpzc{E}$ is a set of effects \textit{eff} $\in \mathpzc{E}$ each of the form $f := v$  with $f \in \mathpzc{V}$ and $v \in Dom(f)$.
        \item $\mathpzc{M}$ is a (possibly empty) set of motion constraints of the form $\langle r, q_S, q_G, \mathpzc{O} \rangle$, where $r \in \mathpzc{R}$ is the agent performing \textit{a}, $q_S \in \mathpzc{C}_r$ is its start configuration, $q_G \in \mathpzc{C}_r$ is the target configuration, and $\mathpzc{O} \subseteq 2^{\mathpzc{R}\times\mathpzc{C}}$ is a function associating the other movable agents, which \textit{r} must avoid, to the configurations they currently occupy.
    \end{itemize}

    \item $\mathpzc{G}$ is the goal condition, represented as a Boolean combination of atoms of the form $f = v$, with $f \in \mathpzc{V}$ and $v \in Dom(f)$.
\end{itemize}
\end{definition}

Focusing on the semantics of the problem, a state $S$ is a total assignment of values to the fluents such that $S(f) \in Dom(f)$ for all $f \in \mathpzc{V}$. An action is applicable in a state $S$ if its preconditions are satisfied by substituting each fluent $f$ appearing in the Boolean combination with $S(f)$ and if all the motion constraints $\mathpzc{M}$ are satisfiable.

A motion constraint $\langle r, q_S, q_G, \mathpzc{O} \rangle$ is satisfied if there exists a collision-free path $\tau$: [0,1] $\rightarrow$ $\mathpzc{C}_r$ that moves $r$ from $\tau(0) = q_S$ to $\tau(1) = q_G$. $\tau$ must be compliant with the motion model $\mathpzc{U}_r$, must reside in $\mathpzc{W}_f$, i.e., $\forall t \in [0, 1] . \: occ(r, \tau(t)) \subseteq \mathpzc{W}_f$, and must be collision-free with the obstacles listed on $O$, i.e., $\forall t \in [0, 1] . \: \forall \langle r', q' \rangle \in O . \: occ(r, \tau(t)) \cap occ(r', q') = \emptyset$.

The successor of $S$, once applied $a = \langle \mathpzc{P}, \mathpzc{E}, \mathpzc{M} \rangle$, is $a(S)$ where:
\begin{equation*}
    a(S)(f) =
    \begin{cases}
        v & \mbox{if } \langle f:=v \rangle \in \mathpzc{E} \\
        S(f) & \mbox{otherwise}
    \end{cases}
\end{equation*}

The plan $\pi$ solving $\psi$ is a sequence $\langle a_0, \ldots, a_n \rangle$ of actions such that $a_0$ is applicable in \textit{I}, each action $a_i$ is applicable in $a_{i-1}(a_{i-2}(\cdots(a_0(I))))$, and the final state 
satisfies $\mathpzc{G}$.


In our example, $r$ moves in a deterministic and fully observable 2D map where fixed obstacles are the walls. Thus, $\mathpzc{W}_f$ encompasses all the points on the map not occupied by the walls. The robot moves according to a ReedsShepp-type motion model~\cite{pjm/1102645450}, i.e., its configuration has the form (\textit{x}, \textit{y}, $\theta$), where (\textit{x}, \textit{y}) are Cartesian coordinates and $\theta$ is the orientation angle. A motion constraint $\langle r, q_S, q_G, \mathpzc{O} \rangle$ is satisfied if we find a path that connects $q_S$ to $q_G$ while avoiding the doors in $O$, where a door is a movable object that changes its configuration from closed to open when pressing its open button.
%

Definition \ref{def:tamp} is a ground formalization of the TAMP problem we tackle. For the sake of brevity, we only formalize the syntax and semantics of the ground representation. In practical modeling, we adopt a lifted representation, as is customary in the planning community. Our peculiarity is to consider movable agents and configurations of interest as objects of the problem, allowing fluents to have subsets of configurations as domains. This is useful for specifying goals for the agents and expressions evaluating movable agents or configurations.
If $e_r$ is an expression evaluating a movable agent and $e_q$ is an expression evaluating a configuration, a motion constraint will have the form $\tuple{e_r, e_{q_S}, e_{q_G}, \mathpzc{O_l}}$, with $O_l$ a set of pairs of the form $\tuple{e_r, e_q}$. The semantics is given by grounding: we assess the expressions within the lifted motion constraint in the state where the action starts, and we obtain the ground motion constraint of Definition \ref{def:tamp}.

\section{Meta-Engine Framework}\label{sec:meta_engine}

To effectively and efficiently solve the TAMP problem $\psi$, we developed a meta-engine framework that allows to interleave an off-the-shelf task planner $\xi$ and an off-the-shelf motion planner $\rho$, provided as inputs. The basic idea of the approach is to invoke the task planner on the planning problem obtained by disregarding all the motion constraints of every action to generate a candidate plan. The candidate plan is then checked to ensure all the motion constraints of the involved actions are realizable. If this is the case, then the plan is returned, otherwise we extract information from the search space of the motion planner for the motion constraint that is not realizable; this information is then used to refine the task problem and we restart the task planner to find a new candidate plan. In this section, we detail this general schema and we explain how the refinement is computed.

\begin{algorithm}[!t]
\footnotesize
\caption{Our Meta-Engine Framework}\label{alg:meta-engine}
\begin{algorithmic}[1]
\Procedure{MetaSolve}{$\psi$, $\xi$, $\rho$, $t_\rho$}
    \State $\gamma \gets \emptyset$ \Comment{Cache for successful motion constraints}
    \State $\psi' \gets \psi$ \Comment{$\psi'$ is the abstracted problem}
    \While{\texttt{True}}
        \State $\pi$ $\leftarrow$ $\xi$.\verb|solve|($\psi'$) \Comment{Call task planner on $\psi'$} \label{alg:meta-solve-func}
        \If{$\pi \neq \emptyset$}
            \State $\nu$, $\mu$ $\leftarrow$ \Call{CheckPlanMotions}{$\rho$, $\gamma$, $\pi$, $t_{\rho}$} \label{alg:meta-check-plan}
            \If{$\nu$}
                \State \Return $\langle \pi, \gamma \rangle$ \Comment{$\pi$ is a valid plan, $\gamma$ has the paths} \label{alg:meta-return-plan}
            \Else
                \State $\psi' \gets $ \Call{RefineProblem}{$\psi'$, $\mu$} \label{alg:refine_psi}
            \EndIf
        \Else
            \State $t_{\rho}$ $\leftarrow$ $t_{\rho} * 2$ \Comment{Increase motion planner timeout} \label{alg:meta-double-time}
            \State $\psi' \gets \psi$ \Comment{Reset the refinements}
        \EndIf
    \EndWhile
\EndProcedure
\end{algorithmic}
\end{algorithm}

\begin{algorithm}[!t]
\footnotesize
\caption{Checking motion constraints in a given plan}\label{alg:check_plan}
\begin{algorithmic}[1]
\Procedure{CheckPlanMotions}{$\rho$, $\gamma$, $\pi$, $t_{\rho}$}
    \State $\nu$ $\leftarrow$ \verb|True| \Comment{Final validity of the plan}
    \State $\mu \leftarrow \emptyset$ \Comment{Set of unsat constraints and learned info}
    \ForEach{ $a = \langle \mathpzc{P}, \mathpzc{E}, \mathpzc{M} \rangle \in \pi$}
        \ForEach{$c = \tuple{r, q_S, q_G, \mathpzc{O}} \in \mathpzc{M}$}
            \If{$\not \exists \langle c, \tau \rangle \in \gamma$} \Comment{$\gamma$ stores past solutions}
                \State $\tau$, $\langle \sigma, \omega\rangle$ $\leftarrow$ $\rho$.\verb|check|($c$, $t_{\rho}$) \label{alg:check-func}
                \If{$\tau \not = \emptyset$} \Comment{If a path $\tau$ is found}
                    \State $\gamma \gets \gamma \cup \{\langle c, \tau\rangle \}$ \Comment{Save path $\tau$ in $\gamma$} \label{alg:check-path}
                \Else \Comment{If no path was found}
                    \State $\nu \gets $ \verb|False| \Comment{$\pi$ cannot be validated}
                    \State $\mu \leftarrow \mu \cup \{\langle r, q_S, \sigma, \omega\rangle\}$ \Comment{Learn info} \label{alg:check-return-refinements}
                \EndIf
            \EndIf
        \EndFor
    \EndFor
    \State \Return $\nu$, $\mu$
\EndProcedure
\end{algorithmic}
\end{algorithm}

Algorithm~\ref{alg:meta-engine} reports the pseudo-code of the meta-engine. The task planner $\xi$ searches for a plan $\pi$  that is valid for the problem $\psi$ while disregarding the motion constraints (line~\ref{alg:meta-solve-func}). By excluding the motion aspect, the problem is reduced to a traditional task-planning problem.
If a valid plan is found, the function \textsc{CheckPlanMotions} checks all the motion constraints of all the actions involved in the plan (line~\ref{alg:meta-check-plan}). Since many motion planning algorithms are sample-based and do not guarantee termination if a path does not exist, we set a timeout $t_{\rho}$ to each invocation of the motion planner. The algorithm keeps a cache $\gamma$ which stores each motion constraint successfully checked and its trajectory $\tau$. If all the motion constraints of all the actions of $\pi$ are found to be realizable by the motion planner, then the plan is returned together with  $\gamma$ (line~\ref{alg:meta-return-plan}). If at least one motion constraint cannot be solved, we refine  $\psi'$ (see \textit{Topological Refinements}). If, instead, fails to find a candidate plan, it suggests either the problem is unsolvable or a previous candidate plan was feasible, but the motion planner couldn't find a path in time. Thus, we double the timeout $t_{\rho}$ of the motion planner, reset our refinements, and restart the algorithm (line~\ref{alg:meta-double-time}). Note that $\gamma$ is not reset, preserving any valid motion plan and improving the efficiency of the algorithm itself.

In Algorithm~\ref{alg:check_plan}, \textsc{CheckPlanMotions} generates a boolean value $\nu$ which is True if the plan is successfully checked. 
Otherwise, the function returns a set $\mu$ of explanations of the form $\tuple{r, q_S, \sigma, \omega}$, where $r \in \mathpzc{R}$, $q_S \in \mathpzc{C}_r$, $\sigma \subseteq \mathpzc{C}_r$ and $\omega \in 2^{\mathpzc{R} \times \mathpzc{C}}$. This means that the motion planner cannot find a path for $r$ in $q_S$ to reach any destination in $\sigma$ with the obstacles $r'$ in $c'$ with $\tuple{r', c'} \in \omega$. 
In the following, we outline how this data is computed from the search space of a sampling-based motion planner and utilized for refinement.



\myparagraph{Topological refinements}
It is crucial for the performance of our technique that the \textsc{CheckPlanMotions} algorithm is capable of providing the explanations $\mu$ for the unsatisfied motion constraints. 
If the constraint $\langle r, q_S, q_G, \mathpzc{O} \rangle$ is infeasible, it means that the target $q_G$ cannot be reached either because it is blocked by fixed obstacles or by movable ones (or that we did not give enough time to the motion planner, but this is handled as discussed above). In the first case, there is simply no plan that solves the high-level task that was assigned to \textit{r}. In the second case, some of the obstacles in $\mathpzc{O}$ prevent $r$ from reaching the target, hence some of them must be moved to find a valid plan.
In our motivating example, this means that some closed door prevents the robotic agent from reaching its final destination.

If the constraint $\langle r, q_S, q_G, \mathpzc{O} \rangle$ is invalid, we find the convex hull 
\begin{equation*}
\mathcal{H}(q_S) = \{\sum_{j=1}^K \lambda_j p_j | \bigwedge_{j=1}^K \lambda_j \geq 0 \wedge \sum_{j=1}^K \lambda_j = 1\}    
\end{equation*}
of the points \{$p_1$, \dots, $p_k$\} sampled by the motion planner from $q_S$. 

Let $X$ be the set \{$q_1$, \dots, $q_m$\} $\subseteq \mathpzc{C}_r$ of \textit{interesting} configurations that the agent may assume, i.e., the motion constraints' configurations involving $r$ for the ground case or the objects of type \textit{Configuration} for the lifted case. We check which configurations yield an occupancy that does not belong to $\mathcal{H}(q_S)$. The idea is that $\mathcal{H}(q_S)$ is an approximation of the positions that the agent can reach and we want to compute the set of interesting locations that are unreachable from the specified starting configuration $q_S$. We call the resulting set $\sigma$ and we define it formally as $\{q \in X \mid occ(r, q) \not \subseteq \mathcal{H}(q_S)\}$. 

The second element of the explanation concerns the blocking movable obstacles. Not all the obstacles in O block the agent from reaching its goal, hence we isolate the obstacles that prevent the motion planner from computing a feasible path connecting $q_S$ to $\sigma$. We call this set $\omega \subseteq \mathpzc{O}$. This set can be efficiently computed by keeping track of the collisions analyzed by the motion planner: if a collision happens in a point $p \in occ(r', q')$ with $\tuple{r',q'} \in \mathpzc{O}$, we add the element $\tuple{r',q'}$ to $\omega$. 
The intuition is that obstacles we do not collide with do not hinder finding a valid plan, offering no useful information for pruning the task planner's search space. Hence, they can be omitted.

\textsc{CheckPlanMotions} collects all the conflicts in $\mu$ and uses this data to refine the problem (line~\ref{alg:refine_psi}). The idea is to prevent the task planner from using actions that are not feasible because of the explanations in $\mu$. We present here two refinements, one for the grounded problem of Definition~\ref{def:tamp} and a more practical one for the lifted case.

In the grounded refinement, we remove any actions with motion constraints that conflict with explanations in $\mu$, thereby refining the set of actions. Formally, given $\psi = \tuple{\mathpzc{R}, \mathpzc{W}, \mathpzc{C}, \mathpzc{U}, \mathpzc{V}, \mathpzc{I}, \mathpzc{A}, \mathpzc{G}}$, we return $\psi' = \tuple{\mathpzc{R}, \mathpzc{W}, \mathpzc{C}, \mathpzc{U}, \mathpzc{V}, \mathpzc{I}, \mathpzc{A}', \mathpzc{G}}$ with $\mathpzc{A'}$ defined as:
\begin{align*}
\{ a = \langle \mathpzc{P}, \mathpzc{E}, \mathpzc{M} & \rangle \in \mathpzc{A} \mid \not \exists m = \tuple{r, q_S, q_G, \mathpzc{O}} \in \mathpzc{M} . \\
& \tuple{r, q_S, \sigma, \omega} \in \mu \land (q_G \in \sigma \lor \omega \subseteq O) \}
\end{align*}
This prevents the execution of actions with known unrealizable constraints (with the given timeout $t_\rho$).

The lifted case is similar, but requires the addition of preconditions to eliminate all the groundings that would conflict with the learned explanations. For each action $a$ in the lifted TAMP problem, we add the following precondition for each lifted motion constraint $m = \tuple{e_r, e_{q_S}, e_{q_G}, \mathpzc{O^l}}$ of $a$ and for each explanation $\tuple{r, q_S, \sigma, \omega} \in \mu$:
\begin{align*}
    & e_r \not = r \vee e_{q_S} \not = q_S \vee \bigwedge_{q \in \sigma} e_{q_G} \not = q \: \vee \\
    & \bigvee_{\tuple{r', c'} \in \omega} \bigwedge_{\tuple{e_r, e_c} \in O^l} ((e_r \not = r') \vee (e_c \not = c'))
\end{align*}
%
which informally means that $m$ of $a$ is consistent with the explanation if any of the following conditions are met: i) $e_r$ does not evaluate to $r$; ii) $e_{q_S}$ does not evaluate to $q_S$; iii) the destination $e_{q_G}$ does not evaluate to any element of $\sigma$; iv) there exists an obstacle in $\omega$ that has a different configuration or doesn't exist in this constraint.

\myparagraph{Theoretical Guarantees}
Many motion planners exist and can be leveraged by our meta-engine. In our case, we exploit sampling-based motion planners, specifically the Rapidly exploring Random Tree (RRT) algorithm~\cite{lavalle1998rapidly} and its Lazy version. Our proposal becomes \textit{probabilistic complete} assuming the task planner is complete, because the probability of finding a solution tends to 1 as the time $t_{\rho}$ given to the motion planner to compute a plan tends to infinity. We also assume that when a motion from $q_S$ to $q_G$ fails, $q_G \in \mu$ at line 7 of Algorithm~\ref{alg:meta-engine} (the destination is always unreachable), preventing to enter an infinite loop as the \textit{interesting} configuration set is finite.

\section{SMT-based Specialization}\label{sec:snt-framework}
\begin{algorithm}[!t]
\footnotesize
\caption{Tampest}\label{alg:tampest}
\begin{algorithmic}[1]
\Procedure{Solve}{$\psi$, $\rho$, $h_{max}$, $t_{\rho}$}
    \State $\gamma \gets \emptyset$
    \While{True}
        \State $\tuple{h, \mu} \gets \tuple{1, \emptyset}$
        \State $\zeta \leftarrow$ \verb|SMTSolver|() \label{alg:tampest-solver}
        \State $\zeta$.\verb|addAssertion|(\verb|initialStep|($\psi$)) \label{alg:tampest-add-initial-state}
        \While{$h \leq h_{max}$}
            \State $f$, $l$ $\leftarrow$ \verb|incrementalStep|($\psi$, $h$) \label{alg:tampest-generate-step}
            \State $\zeta$.\verb|addAssertion|($f$) \label{alg:tampest-add-step}
            \If{$\mu \neq \emptyset$}
            \State $\zeta$.\verb|addAssertion|(\verb|getLemmas|($\mu$, $h$)) \label{alg:tampest-cached}
            \EndIf
            \State $\zeta$.\verb|push|()
            \State $\zeta$.\verb|addAssertion|($l$) \label{alg:tampest-add-goal}
            \While{$\zeta$.solve()}
                \State $\pi$ $\leftarrow$ \verb|getPlan|($\zeta$.\verb|getModel|()) \label{alg:tampest-get-plan}
                \State $\nu$, $\mu'$ $\leftarrow$ \Call{CheckPlanMotions}{$\rho$, $\gamma$, $\pi$, $t_{\rho}$} \label{alg:tampest-check-plan}
                \If{$\nu$}
                    \State \Return $\tuple{\pi, \gamma}$ \label{alg:tampest-return-plan}
                \Else
                    \State $\zeta$.\verb|pop|() \label{alg:tampest-pop}
                        \ForEach{$i \in \{1, \ldots, h\}$} \label{alg:tampest-get-conflict-lemma-start}
                            \State $\zeta$.\verb|addAssertion|(\verb|getLemmas|($\mu'$, $i$))
                    \EndFor \label{alg:tampest-get-conflict-lemma-end}
                    \State $\mu \leftarrow \mu \cup \mu'$
                    \State $\zeta$.\verb|push|()
                    \State $\zeta$.\verb|addAssertion|($l$) \label{alg:tampest-last}
                \EndIf
            \EndWhile
            \State $\zeta$.\verb|pop|()
            \State $h \leftarrow h+1$
        \EndWhile
        \State $t_{\rho} \leftarrow t_{\rho}*2$ \label{alg:tampest-double-time}
    \EndWhile
\EndProcedure
\end{algorithmic}
\end{algorithm}

We tailored our framework to leverage the incremental solution capabilities of SMT-based solvers. Such solvers maintain a stack of constraints (called assertions), enabling efficient repeated satisfiability checks as constraints are pushed onto or popped from the constraint stack. This feature eliminates the need for restarting the planning routine upon failure to find a valid plan, enhancing overall scalability.

Our approach is called \textsc{Tampest} and it iterates between task and motion planning while progressively increasing the search depth until finding a valid plan or reaching the maximum step horizon $h_{max}$. 

As shown in Algorithm~\ref{alg:tampest}, the general schema is that of the meta-engine in Algorithm~\ref{alg:meta-engine}, with the outer while loop serving for the refinement of the motion planner timeout $t_\rho$, the learned explanations $\mu$, and the horizon $h$. The inner loop is the focal point of the approach.
We encode the task part of $\psi$ as an SMT planning problem, analogously to many SATPlan-like approaches \cite{satplan,rintanen}, and we add to $\zeta$ the assertions relative to the initial state, which hold at step 0 (line~\ref{alg:tampest-add-initial-state}).
At each step $h \leq h_{max}$, we generate and add the assertions $f$ and $l$ (lines~\ref{alg:tampest-add-step}-\ref{alg:tampest-add-goal}). As in~\cite{Dantam2020}, $f$ asserts that a selected action implies its preconditions and effects, the state remains the same unless changed by an action effect, and only one subset of non-mutex actions is taken at time. Assertion \textit{l}, instead, characterizes the goal.
$\zeta$ searches for a valid plan $\pi$, that means finding a satisfying assignment for the asserted logical formulae (line~\ref{alg:tampest-get-plan}).
If a model exists, we check the motion feasibility of $\pi$ via \textsc{CheckPlanMotions}, possibly exploiting the cached information (line~\ref{alg:tampest-check-plan}).
If all constraints are satisfied, we return the plan and the paths (line~\ref{alg:tampest-return-plan}).
Otherwise, we pop the solver and add the logical lemmas representing the topological refinements $\mu'$. We use the same logical formulation used for the lifted refinement in the meta-engine encoding the preconditions as an SMT formula instantiated at all the symbolic times $i \in \{1, \ldots, h\}$. 
Once this data is added, we push the solver, re-add the goal, and try to find a solution again (lines~\ref{alg:tampest-pop}:\ref{alg:tampest-last}).
Every time we enlarge the encoding bound, we permanently add the lemmas for all the explanations in $\mu$ at $h$, ensuring their validity across all encoding steps (line~\ref{alg:tampest-cached}).

\section{Modeling and Benchmarking}\label{sec:modeling_and_benchnmarking}
\begin{figure*}[t]
     \centering
     \begin{subfigure}[b]{0.33\textwidth}
         \centering
         \includegraphics[width=0.9\textwidth]{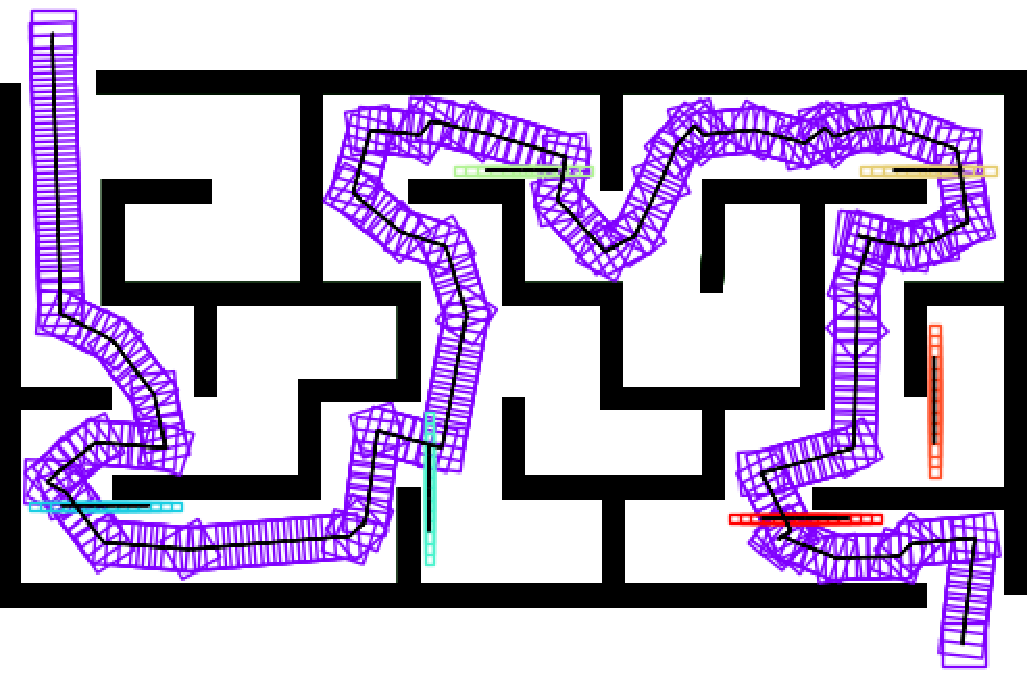}
         \caption{}
         \label{subfig:maze_domain}
     \end{subfigure}
     \hfill
     \begin{subfigure}[b]{0.32\textwidth}
         \centering
         \includegraphics[width=0.9\textwidth]{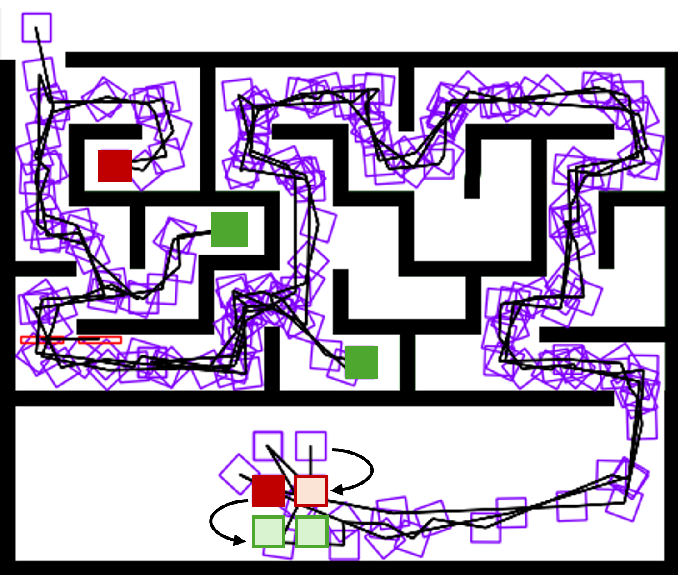}
         \caption{}
         \label{subfig:delivery_domain}
     \end{subfigure}
     \hfill
     \begin{subfigure}[b]{0.32\textwidth}
         \centering
         \includegraphics[width=0.865\textwidth]{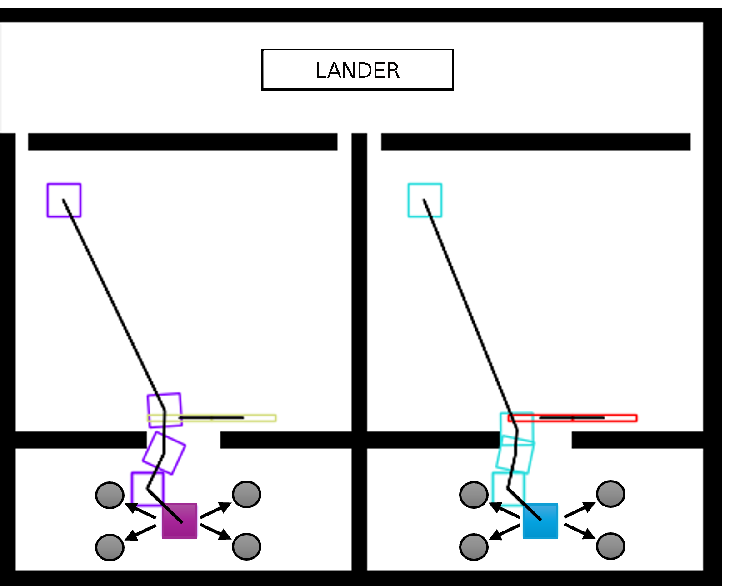}
         \caption{}
         \label{subfig:rovers_domain}
     \end{subfigure}
     \vspace{0.25cm}
    \caption{2D setups of our benchmarks. \subref{subfig:maze_domain} \textit{Maze} with $N = 5$ blocking doors. \subref{subfig:delivery_domain} \textit{Delivery} with $N=1$ door and $M=4$ parcels, one at its delivery station. \subref{subfig:rovers_domain} \textit{Rovers} with $N=2$ rovers collecting a rock and a soil sample, and taking images of $M=4$ objectives around the samples.}
    \vspace{0.1cm}
    \label{fig:benchmarks}
\end{figure*}

Besides formulating the TAMP problem of Definition~\ref{def:tamp} and defining suitable TAMP solvers, we developed a comprehensive open-source framework for modeling and benchmarking these problems. An overview of the key components of this implementation follows, along with a description of the benchmark suite we designed. 

UP\footnoteref{up} is an open-source, planner-agnostic planning library that collects planning tools and algorithms to model, manipulate, and solve classical, numerical, temporal, and other complex tasks, such as multi-agent assignments.
To enable the modeling of TAMP problems, we extended the TAMP modeling of the UP adding obstacle avoidance. 
Besides preconditions and effects, motion actions include motion constraints of the form \textit{path(r, $q_S$, [$q_G$], \{o : $q_o$ $\forall$ o $\in \mathcal{O}$\})}, i.e., there $\exists \pi:[q_S, [q_G]] \rightarrow \mathcal{C}_r$ for $r \in \mathcal{R}$ and \{\textit{o : $q_o$ $\forall$ o $\in \mathcal{O}$}\}, as in Definition~\ref{def:tamp}.
Non-fixed objects are defined as \textit{Movable Objects} with a geometric and motion model. Their configurations are \textit{Configuration Objects} with a value in the form provided by the motion model of the agent (e.g., (\textit{x}, \textit{y}, \textit{yaw}) in SE(2)). The workspace is an \textit{Occupancy Map} collecting all useful data for motion planning and collision avoidance with fixed obstacles, such as the 2D image or 3D mesh of the operating environment and its reference system.
We allow fluents that accept as input a \textit{Movable Object} and output its current \textit{Configuration Object} within the \textit{Occupancy Map.}
As for all the tools of the UP library, this extension is independent of the planning language and planner available to define and solve this problem.

With this extension, we offer a set of benchmarks that task robotic agents with Navigating Among Movable Obstacles (NAMO)~\cite{Stilman}, i.e., moving through a workspace while removing or avoiding movable obstacles.
As in~\cite{8411475}, we assume the search space is i) \textit{geometric}: motion planning focuses only on finding feasible object poses based on the geometric constraints of the world; ii) \textit{fully observable}: the initial state is completely known both geometrically 
and semantically
; iii) \textit{deterministic}: world state changes exclusively result from planned actions, and object motions precisely adhere to the motion planner's output. We consider the following evaluation criteria:
\begin{itemize}
    \item \textbf{Infeasible task actions}. Some task actions are impossible due to the lack of corresponding feasible motion plans caused by obstructing obstacles.
    \item \textbf{Large task spaces}. The task planning problem requires substantial search effort.
    \item \textbf{Motion/Task Trade-off}. The problem can be solved with fewer steps if the right obstacles are moved.
    \item \textbf{Non-monotonicity}. Some objects need to be moved more than once for achieving the goal.
    \item \textbf{Non-geometric actions}. Some actions, like perception, change the discrete state but not the robot configuration.
\end{itemize}
The description of our benchmarks follows.
For each domain, we offer a comprehensive setup, ensuring faithful replication in both 2D and 3D environments. This approach guarantees reliable assessment of solver performance, even within complex search spaces.
In 2D scenarios, movable objects are polygons and the robot navigates using a ReedsShepp path within a black-and-white map, where black represents areas occupied by fixed obstacles. In 3D, objects are 3D rigid bodies and move according to an SE(3) motion model. Due to limited space, we'll discuss only the benchmarks deemed paradigmatic according to the outlined evaluation criteria (see Table~\ref{tab:benchmarks_criteria}):
\begin{table}[t]
    \centering
    \begin{tabular}{|l|c| c|c|c|}
        \hline
        \textbf{Criteria}           &  \textbf{Doors}       &  \textbf{Maze}        & \textbf{Delivery}     & \textbf{Rover}     \\ \hline
        Infeasible task actions     &  x                    &  x                    &   x                   & x                     \\ \hline
        Large task spaces           &  x                    &  x                    &   x                   & x                     \\ \hline
        Motion/Task trade-off       &                       &  x                    &   x                   & x                     \\ \hline
        Non-monotonicity            &                       &                       &   x                   & x                     \\ \hline
        Non-geometric actions       &                       &                       &                       & x                     \\ \hline
    \end{tabular}
    \vspace{0.2cm}
    \caption{Criteria evaluated by each benchmark problem.}
    \label{tab:benchmarks_criteria}
\end{table}

\begin{itemize}
    \item \textbf{Doors}. 
    One robot needs to navigate through \textit{N} initially closed doors to reach a final destination, using the  \{\textit{move}, \textit{open}\} action set (see Figure~\ref{fig:doors_domain}). \textit{move} enables the robot to navigate from a start to a goal location and incorporates a motion constraint that avoids collisions with movable and static obstacles (doors and walls). \textit{open} allows the robot to open a door when positioned in front of it, like pushing a button. Once the button is pushed, the door configuration changes instantaneously from closed to open. \textit{M} extra locations are randomly sampled in the free space. All locations are connected from a task planning perspective, but the additional ones don't aid in achieving the goal; they merely expand the task space. Thus, even if the problem is simple, the optimal plan contains $2N+1$ steps 
    while the worst-case scenario needs $2N+M+1$ steps to take the robot from start to goal while opening all the doors and visiting all extra locations (\textit{large task space}). Closed doors make some locations unreachable (\textit{infeasible task actions}).

    \item \textbf{Maze}. A robot must navigate out of a maze while visiting \textit{M} points randomly distributed within it (see Figure~\ref{subfig:maze_domain}). \textit{N} doors block various passages, not all leading to exit or target locations. Their motion model requires the motion planner to compute opening paths. Actions are \{\textit{move}, \textit{open}\}.
    Again, we are exploring a \textit{large task space} equipped with \textit{infeasible task actions}. Moreover, we should find a good \textit{motion/task trade-off} to efficiently solve the problem: while opening all doors and reaching the assigned targets is valid, only opening necessary doors yields efficiency.

    \item \textbf{Delivery}. Inspired by the \textit{delivery} domain of IPC, \textit{Maze} locations become parcels with no geometry and motion model. They are distinguished by colors and must be arranged into rows by color, each row delivered before the next (see Figure~\ref{subfig:delivery_domain}). Actions are \{\textit{move}, \textit{open}, \textit{load}, \textit{unload}\}, where \textit{load} involves collecting a parcel and placing it atop an agent. \textit{unload} enables the agent to remove an item from its cargo and deposit it at a specified location (\textit{large task space}). The robot has a fixed capacity (\textit{numerical problem}), and can unload packages only when positioned in front of the unloading location, though some parcels are already at their stations. \textit{N} doors block \textit{N} passages, some of which are useful to reach the unloading area (\textit{unfeasible task actions}).
    The layout of the unloading area and the presence of obstructing doors influence the \textit{motion/task trade-off}. Parcels initially at unloading stations enable assessment of \textit{non-monotonicity}: if a parcel blocks the unloading of other items, it must be temporally relocated.

    \item \textbf{Rovers}. We reproduce the \textit{rover} domain of IPC to demonstrate the generality of our approach (see Figure~\ref{subfig:rovers_domain}).  \textit{N} rovers must collect rock and soil samples, separated from each robot by a door. Then, they must calibrate their cameras, photograph \textit{M} objectives located around each sample without occlusions, and send the results back to a lander. Due to obstacles that limit the reachability of parts of the workspace, one rover must be utilized for each sample and the objectives around it.
    Actions are \{\textit{move}, \textit{open}, \textit{calibrate}, \textit{sample rock}, \textit{sample soil}, \textit{send analysis}, \textit{drop}, \textit{take image}, \textit{send image}\}, and some of them change only the discrete state and not the configuration space (\textit{non-geometric actions}).
\end{itemize}


\section{Related Work}\label{sec:state-of-the-art}
Many planners exist that combine symbolic and geometric search.
As an example, the aSyMov planner~\cite{cambon2004robot,cambon2009hybrid}
interleaves a FF-based task planner with lazily-expanded roadmaps. However, this approach is impractical when action plans are valid in the symbolic space but infeasible in the geometric one.
To address this issue, many approaches have been developed over the years. For instance, Dornhege~\emph{et al.}~\cite{dornhege2012semantic} add semantic attachments to the definition of the task, and they call the motion planner after each action to check both its geometric and semantic feasibility.
Other strategies, as discussed in~\cite{kaelbling2011hierarchical, srivastava2014combined, FFRob, Toussaint, 7354287}, are tailored to specific classes of manipulation problems, limiting their adaptability to new domains, such as those introduced in this paper, without significant engineering effort. They lack a modular, domain-agnostic problem description language with clear semantics.
In this regard, PDDLStream integrates symbolic planners and black-box samplers by extending Planning Domain Definition Language (PDDL)~\cite{pddl} with streams: declarative specifications of sampling procedures that link in a black-box way the symbolic representation of constraints with their sample-based counterparts. In TAMP, they are used to map the existence of collision-free paths with the functions checking their validity. 
Our formulation is less general as it is tailored specifically towards TAMP problems. However, this targeted approach allows us to exploit the motion planner's output to prune large regions of the task search space, significantly reducing the computational overhead.

Indeed, calling the motion planner after each symbolic call is time-consuming, particularly when dealing with geometrically unfeasible states. To enhance efficiency, the geometric search is typically limited to candidate symbolic plans.
Srivastava~\emph{et al.}~\cite{srivastava2014combined}, for example, interface a task planner with an optimization-based motion planner and use a heuristic to remove occluding objects.
Dantam~\emph{et al.}.~\cite{dantam2016incremental} propose TMKit: an incremental SMT solver that incrementally generates symbolic plans and call the motion planner for validation.
They all suffer from long processing time, solve problems consisting of a limited number of actions and, given their focus on manipulation tasks, handle a limited quantity of manipulable objects. Some approaches exist that tries to overcome these limitations. 
Similar to TMKit, our SMT specialization employs an incremental approach to generate a valid symbolic plan. Initially, it assumes the validity of all motion actions within the plan. Once a task plan is established, it invokes the motion planner to verify feasibility. For any unfeasible motion action, we generate topological refinements on the geometric space. These refinements are leveraged at the task level, enhancing efficiency and allowing for plans with many actions.

\begin{figure}[!t]
    \centering
    \includegraphics[width=0.45\textwidth]{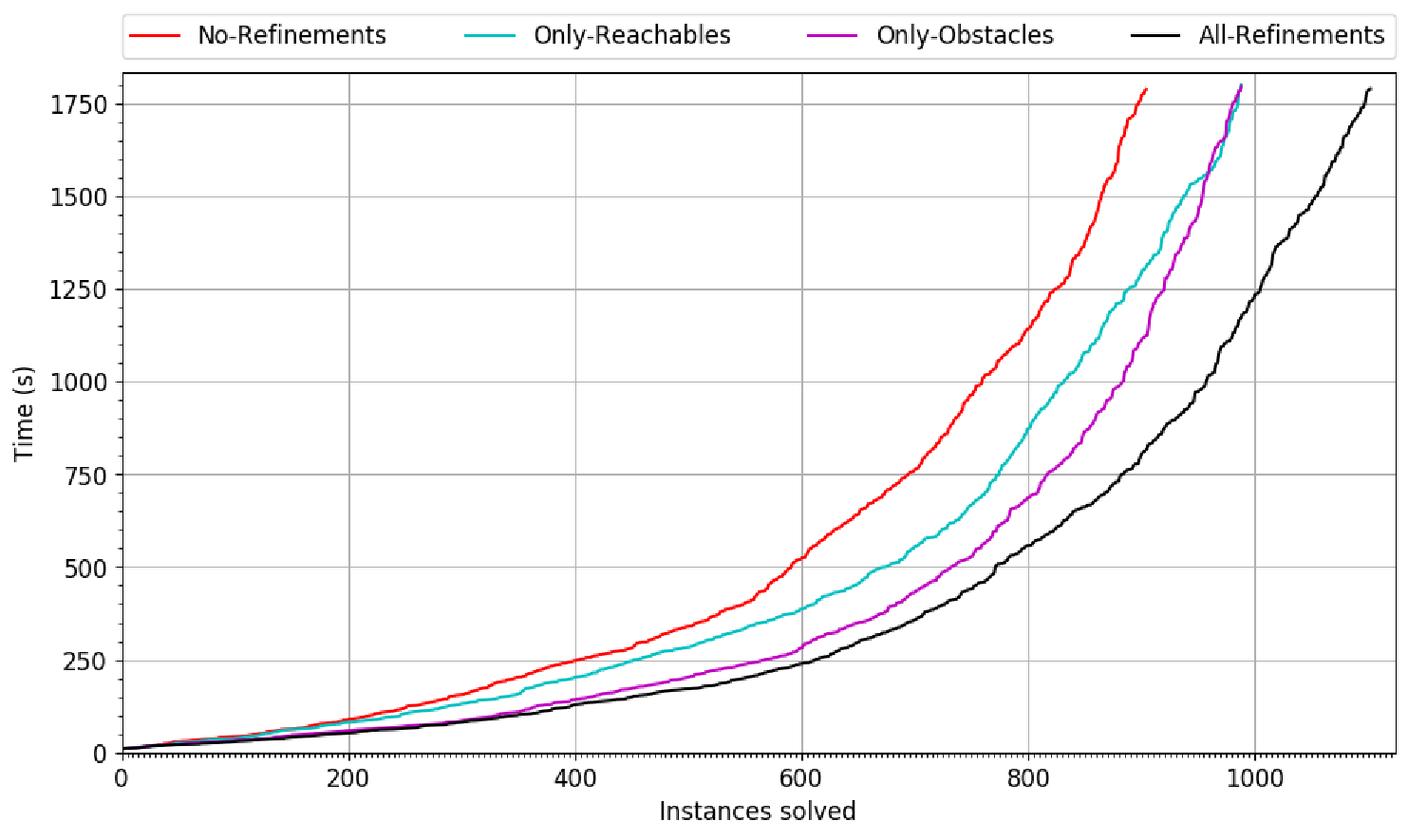}
    \caption{Overall performance on all benchmark instances and all planners when exploiting different topological refinements.}
    \vspace{0.5cm}
    \label{fig:topological-refinements}
\end{figure}

\section{Experimental Evaluation}\label{sec:experimental_evaluation}
\begin{figure*}[!t]
     \centering
     \begin{subfigure}[b]{0.3\textwidth}
         \centering
         \includegraphics[width=\textwidth]{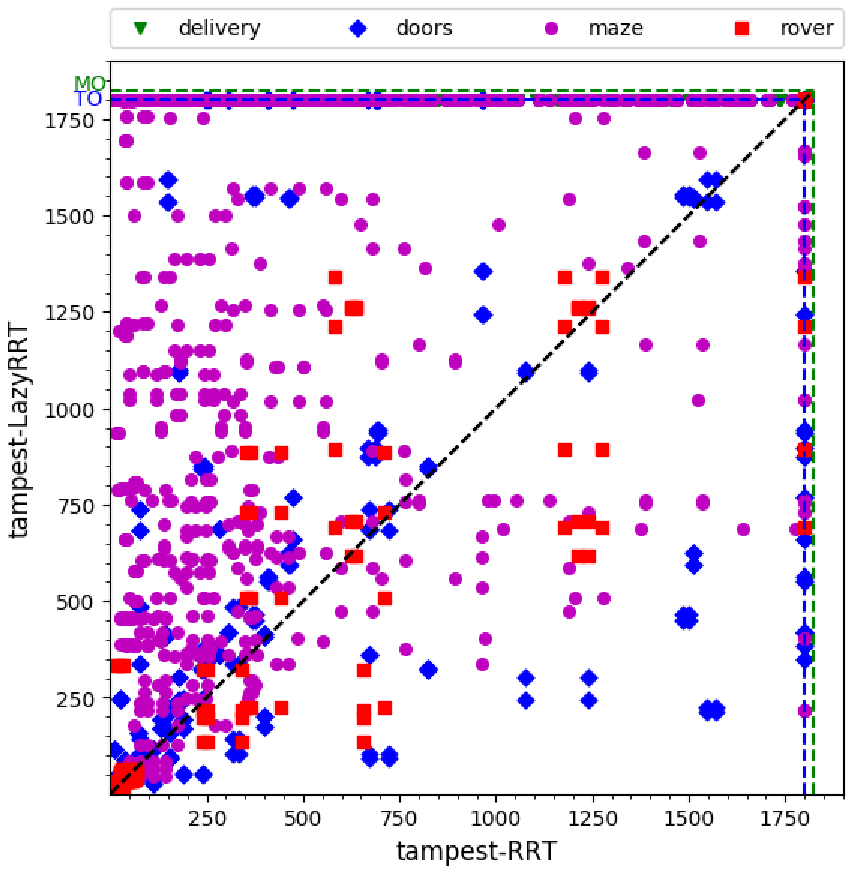}
         \caption{}\label{fig:rrt-lazyrrt}
     \end{subfigure}
     \begin{subfigure}[b]{0.3\textwidth}
         \centering
         \includegraphics[width=\textwidth]{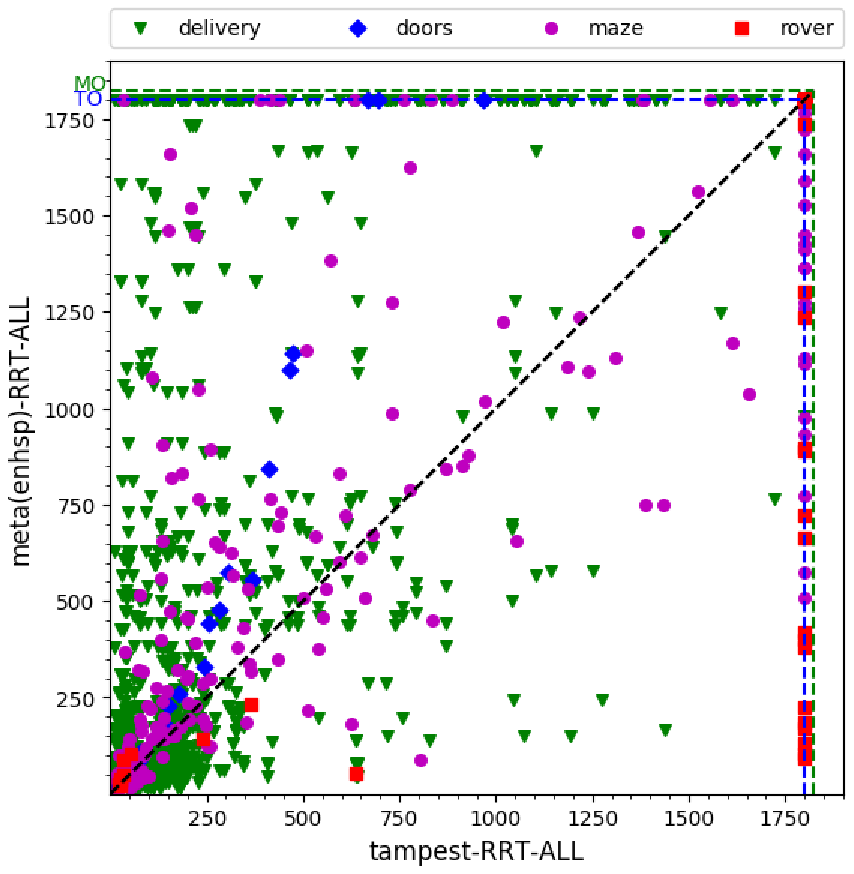}
         \caption{}\label{fig:tampest-enhsp}
     \end{subfigure}
     \begin{subfigure}[b]{0.3\textwidth}
         \centering
         \includegraphics[width=\textwidth]{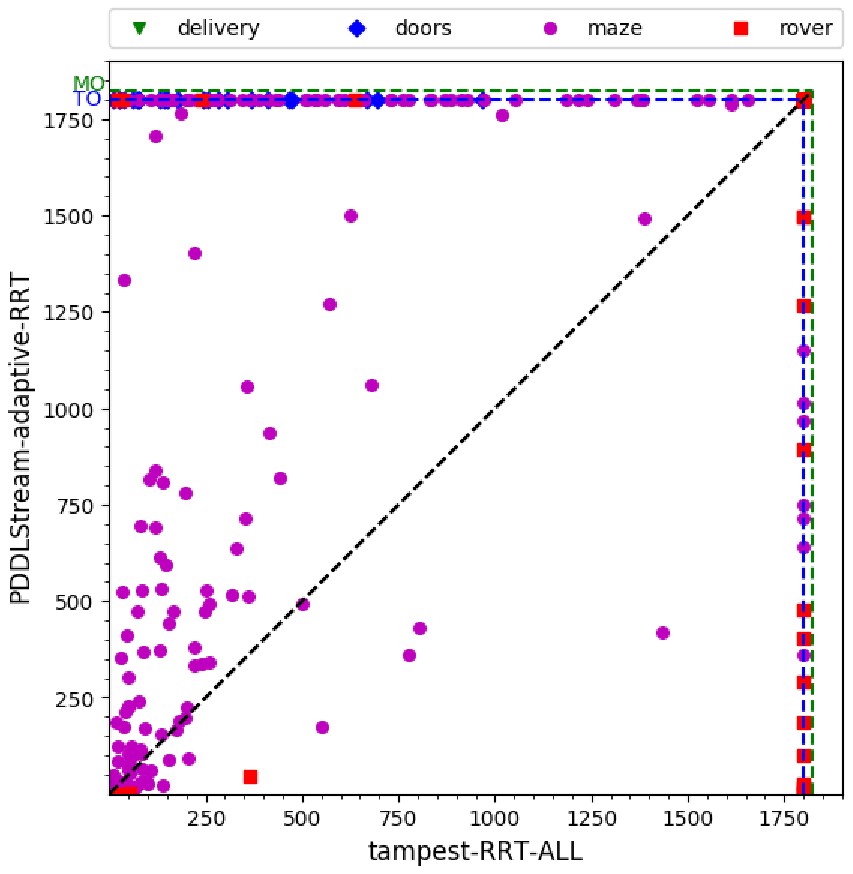}
         \caption{}\label{fig:tampest-adaptive}
     \end{subfigure}
     \vspace{0.24cm}
        \caption{Comparing (a) \textsc{RRT} vs. \textsc{LazyRRT} when using \textsc{Tampest}, regardless of the refinement exploited. (b) \textsc{Tampest} vs. \textit{Meta}(\textsc{ENHPS}), both with \textsc{RRT} and \textit{All-Refinements}. (c) \textsc{Tampest} vs. \textsc{PDDLStream}-\textit{adaptive}, both with \textsc{RRT} and \textsc{Tampest} with \textit{All-Refinements}.}
    \vspace{0.2cm}
        \label{fig:cactus}
\end{figure*}

\begin{table*}[t]
\centering
\begin{tabular}{|l|cc|cc|cc|cc|}
\hline
\textbf{Planner}                                     & \multicolumn{2}{c|}{\textbf{Doors} (tot. 24)} & \multicolumn{2}{l|}{\textbf{Maze} (tot. 220)} & \multicolumn{2}{l|}{\textbf{Delivery} (tot. 525)} & \multicolumn{2}{l|}{\textbf{Rover} (tot. 50)} \\ \hline
\textsc{PDDLStream}-\textit{binding}       & \multicolumn{2}{c|}{2}      & \multicolumn{2}{c|}{17}     & \multicolumn{2}{c|}{-}         & \multicolumn{2}{c|}{0}      \\ \hline
\textsc{PDDLStream}-\textit{focused}        & \multicolumn{2}{c|}{0}      & \multicolumn{2}{c|}{17}     & \multicolumn{2}{c|}{-}         & \multicolumn{2}{c|}{0}      \\ \hline
\textsc{PDDLStream}-\textit{incremental}    & \multicolumn{2}{c|}{6}      & \multicolumn{2}{c|}{43}     & \multicolumn{2}{c|}{-}         & \multicolumn{2}{c|}{1}      \\ \hline
\textsc{PDDLStream}-\textit{adaptive}       & \multicolumn{2}{c|}{1}      & \multicolumn{2}{c|}{65}     & \multicolumn{2}{c|}{-}         & \multicolumn{2}{c|}{1}      \\ \hline
\textit{Meta}(\textsc{Fast-Downward})                      & \multicolumn{1}{c|}{4}  & 3 & \multicolumn{1}{c|}{52}  & 67 & \multicolumn{2}{c|}{-}  & \multicolumn{1}{l|}{7}  & 9 \\ \hline
\textit{Meta}(\textsc{Tamer})                              & \multicolumn{1}{c|}{4}  &  7 & \multicolumn{1}{c|}{51}  & 56 & \multicolumn{1}{l|}{376}    &  376  & \multicolumn{1}{l|}{12}  & 11  \\ \hline
\textit{Meta}(\textsc{ENHSP})                              & \multicolumn{1}{c|}{13}  & 21  & \multicolumn{1}{c|}{121}  & 154 & \multicolumn{1}{l|}{287}    & 287   & \multicolumn{1}{l|}{18}  & \textbf{27}  \\ \hline
\textsc{Tampest}                            & \multicolumn{1}{c|}{17}  & \textbf{24}  & \multicolumn{1}{c|}{126}  & \textbf{164} & \multicolumn{1}{l|}{415}    &  \textbf{422}  & \multicolumn{1}{l|}{12}  & 14  \\ \hline
\end{tabular}
\vspace{0.2cm}
    \caption{Overall performance of all planners on all benchmarks when combined with \textsc{RRT} (left column with \textit{No-Refinement}, right column with \textit{All-Refinements}). All \textsc{PDDLStream} variants are equipped with \textsc{Fast-Downward}, as provided by default by this framework. 
    }
\vspace{0.2cm}
\label{tab:planners}
\end{table*}

In this Section, we present experiments evaluating our meta-engine framework across various task and motion planners. We assess the effectiveness of its SMT-based specialization and quantify improvements from topological refinements. Moreover, we compare our framework with \textsc{PDDLStream}, highlighting our ability to integrate existing solvers and the superior performance of our proposal.
Benchmarks and solvers are available in the supplementary material, to be released upon paper acceptance. Our test cases follows:
\begin{itemize}
    \item \textbf{Doors}. We feature $n_d \in [1, 2, 4, \dots, 10]$ closed doors that must all be open to reach the final destination. Additionally, either 0 ($n_c$ = [(0, 0)]) or 10 extra configurations are randomly distributed in the reachable space ($n_c$ = [(10,0)]), the initially unreachable space ($n_c$ = [(0, 10)]), or equally split between both ($n_c$ = [(5,5)]).
    \item \textbf{Maze}. We increase the complexity of our domain by introducing $n_d \in [1, 2, 3, \dots, 10]$ closed doors within a maze setup, where not all doors block the final destination. The extra-configurations becomes $n_c \in [0, 1, 2, 3, \dots, 10]$ mandatory targets for inspection, randomly located within the maze.
    \item \textbf{Delivery}. We sample $n_d \in [1, 2, 4, \dots, 10]$ closed doors, not all obstructing the target, and $n_r + n_g \in [0, 1, 2, 3, \dots, 8]$ red and green parcels. Colors are randomly sampled among available parcels. Parcels must be delivered in two rows, with at most 4 red parcels placed in the front and 4 green parcels in the back. $d_r \leq 3$ red parcels and $d_g \leq 3$ green parcels are already in their delivery spots, eventually blocking the reachability of the unloading locations behind them, that means $n_c=x=[(n_r, n_g, d_r, d_g)]$. The robot's load capacity $n_l$ ranges from 1 to 4.
    \item \textbf{Rovers}. We involve $2n_d$ robots with $n_d \in [1, 2, 3, 4, 5]$. Each robot analyzes either one soil or one rock sample, each one situated one closed door away from the robot. We design $n_c \in [0, 1, 2, 3, 4]$ objectives to be photographed around each sample.
\end{itemize}
We tested \textit{Maze} and \textit{Rover} domains in both 2D and 3D setups, while \textit{Doors} and \textit{Delivery} tests were limited to their 2D implementations. Indeed, these setups closely resemble those of the former domains.

We instantiate our \textit{Meta-Engine} with \textsc{Fast-Downward}~\cite{fast-downward}, the Expressive Numeric Heuristic Search Planner (\textsc{ENHSP})~\cite{enhsp}, and \textsc{Tamer}~\cite{tamer}, and evaluate their performance compared with \textsc{Tampest} (with $h_{max}$ = 100), where the last three can solve numerical problems such as our \textit{Delivery} domain.
We combine each solver with the \textsc{RRT}~\cite{lavalle1998rapidly} and \textsc{LazyRRT} motion planners (with $t_{\rho}$ = 3s).
In 2D scenarios, we implement an ad-hoc collision checker that verifies the feasibility of a robot's pose by ensuring that its footprint doesn't intersect obstacles. In 3D, we exploit the Flexible Collision Library~\cite{FCL}.
Finally, we study our refinement schema by disabling some of the explanations computed by \textsc{CheckPlanMotions}. We set the topological refinements $\mu' = \{\langle\sigma, \omega\rangle\}$ as follows. \textit{All-Refinements} is the full algorithm as described in the previous sections. \textit{Only-Reachables} assumes $\omega = O$, disabling the analysis of the obstacles with which the agent collided, but retaining the analysis of the unreachable points. \textit{Only-Obstacles} forces $\sigma=\{q_G\}$, retaining the obstacles analysis but disabling the unreachable configurations one, to only remove the target location. \textit{No-Refinements} forces $\sigma=\{q_G\}$ and $\omega = O$, removing only the violated constraint.

Focusing on \textsc{PDDLStream}, we explore its \textit{incremental}, \textit{focused}, \textit{binding}, and \textit{adaptive} variants equipped with \textsc{Fast-Downward}~\cite{FastDownward}, as provided by default. To enable them to solve our benchmarks, 
we convert the motion constraints into streams, mapped with functions that certificates the existence of paths. We employ the same motion planners and collision checkers as before.

We set a global timeout of 1800 s, a memory limit of 10 GB, and ran tests on an Intel Xeon CPU 6226R @2.9GHz.

\myparagraph{Results}
In Figure~\ref{fig:topological-refinements}, we show the impact of leveraging topological refinements across all instances of all domains. The x-axis denotes the number of solved instances, while the y-axis represents computational time. Utilizing \textit{All-Refinements} increases the number of solved instances by roughly 20\% compared to single refinements and 30\% compared to none, also reducing computational time. Some instances have numerous obstacles obstructing large portions of the workspace, highlighting the usefulness of leveraging topological refinements, especially in scenarios with a high number of \textit{infeasible task actions}.

Focusing on the motion planner, \textsc{RRT} outperforms its lazy version, which performs collision checking only at the end. Indeed, our setups feature many obstacles, causing \textsc{LazyRRT} to add a significant number of validation steps during collision checking. Figure~\ref{fig:rrt-lazyrrt} proves this statement when using \textsc{Tampest}, especially for the \textit{Delivery case}, where \textsc{LazyRRT} timeouts in all cases.

In Table~\ref{tab:planners}, we compare planners across all domains, once selected \textsc{RRT}. All \textsc{PDDLStream} variants exhibit lower performance compared to other algorithms, with \textit{adaptive} showing the best results, followed by \textit{incremental}.
Indeed, \textit{incremental} generates all possible streams in advance and then searches for a plan, while \textit{adaptive} first finds a plan and then checks its validity from a motion standpoint. Moreover, it dynamically adjusts its search strategy based on execution progress.
The lower performance of these variants may stem from their use of \textsc{Fast-Downward}, which also affects our \textit{Meta}(\textsc{Fast-Downward})'s results. \textit{Meta}(\textsc{Tamer}) has good performance in the numeric case, while \textit{Meta}(\textsc{ENHSP}) and \textsc{Tampest} demonstrate the highest success rates. This means they can manage \textit{large task spaces} more effectively, achieving a good \textit{trade-off} between motion and task.
In Figure~\ref{fig:tampest-enhsp}, we better compare  the quality and quantity of the solutions proposed by these two algorithms, each equipped with \textsc{RRT} and all topological refinements.
Our proposal performs particularly well in the \textit{Maze} (magenta dots) and \textit{Delivery} (green triangles) domains, i.e., it can face effectively also \textit{non-monotonic} scenarios.
When adding \textit{non-geometric} actions as in the \textit{Rover} domain (red squares), instead, our solver excels with simpler instances, but encounters scalability issues with its SMT component when plans include many actions. As plan size grows, solver performance may decline due to the need to process a larger number of parameters, resulting in longer resolution times.
Finally, in Figure~\ref{fig:tampest-adaptive} we compare \textsc{PDDLStream}'s \textit{adaptive} variant with \textsc{Tampest} (both with \textsc{RRT} and \textsc{Tampest} with all refinements).
The former consistently times out, even when our approach easily finds solutions. This stands out notably in the \textit{Maze} domain (magenta dots). 

\section{Conclusion and Future Work}\label{sec:conclusions}
In this paper, we provided a detailed representation of a multi-agent TAMP scenario with one agent moving at a time and multiple task-dependent obstacles. Our contributions include a general problem formulation and semantic definition, supported by an open-source library for modeling and benchmarking.
We also introduced a novel meta-engine framework for combining off-the-shelf task and motion planners to solve complex scenarios. We proposed using geometric context to generate topological refinements and prune the task planner's search space.
Additionally, we demonstrated how this meta-engine can be adapted for an incremental SMT-based task planner, named \textsc{Tampest}. 
We compared \textsc{Tampest} with existing planners interleaved with sample-based motion planners, with and without topological refinements. SMT's incremental nature accelerates problem resolution, while topological refinements decrease the time required to find a valid plan.  Finally, we integrated PDDLStream enabling direct comparison of solvers on the same input data: \textsc{Tampest} outperforms PDDLStream, especially when using topological refinements.

In future work, we will include metric time and address scenarios with multiple agents moving simultaneously. We will also integrate replanning mechanisms to handle non-determinism.




\end{document}